\documentclass[10pt,twocolumn,letterpaper]{article}

\usepackage{iccv}
\usepackage{times}
\usepackage{epsfig}
\usepackage{graphicx}
\usepackage{amsmath}
\usepackage{amssymb}
\usepackage{booktabs}
\usepackage{bbding}
\usepackage{multirow}
\usepackage{subcaption}
\usepackage{booktabs}
\usepackage{xcolor}
\usepackage[pagebackref=true,breaklinks=true,letterpaper=true,colorlinks,bookmarks=false]{hyperref}

\iccvfinalcopy 

\def\iccvPaperID{****} 

\newcommand{\alg}{EoRaS}

\ificcvfinal\fi

\begin{document}

\title{Rethinking Amodal Video Segmentation from  Learning  Supervised Signals with Object-centric Representation}

\author{Ke Fan\textsuperscript{1,$\dagger$}, 
Jingshi Lei\textsuperscript{1,$\dagger$},
Xuelin Qian\textsuperscript{1,$\ast$},
Miaopeng Yu\textsuperscript{1},
Tianjun Xiao\textsuperscript{2,$\ast$},
Tong He\textsuperscript{2}\\
Zheng Zhang\textsuperscript{2},
Yanwei Fu\textsuperscript{1}
\\
\textsuperscript{1}{Fudan University} \quad
\textsuperscript{2}{Amazon Web Service} \\
{\tt\small \{kfan21,jslei21\}@m.fudan.edu.cn, \{xlqian,mpyu19,yanweifu\}@fudan.edu.cn}\\
{\tt\small \{tianjux, htong, zhaz\}@amazon.com}
\thanks{$\dagger$ co-first authors; $\ast$ corresponding authors. }
}

\makeatletter
\def\thanks#1{\protected@xdef\@thanks{\@thanks
        \protect\footnotetext{#1}}}
\makeatother

\maketitle
\ificcvfinal\thispagestyle{empty}\fi

\begin{abstract}
Video amodal segmentation is a particularly challenging task in computer vision, which requires to deduce the full shape of an object from the visible parts of it.
Recently, some studies have achieved promising performance by using motion flow to integrate information across frames under a self-supervised setting. However, motion flow has a clear limitation by the two factors of moving cameras and object deformation. 
This paper presents a rethinking to previous works. 
We particularly leverage the supervised signals with object-centric representation in \textit{real-world scenarios}.
The underlying idea is the supervision signal of the specific object and the features from different views can mutually benefit the deduction of the full mask in any specific frame. 
We thus propose an Efficient  object-centric Representation amodal Segmentation (EoRaS).
Specially, beyond solely relying on supervision signals, we  design a translation module to project image features into the Bird's-Eye View (BEV), 
which introduces 3D information to improve current feature quality.  
Furthermore, we propose a multi-view fusion layer based temporal module which is equipped with a set of object slots and interacts with features from different views by attention mechanism to fulfill sufficient 
object representation completion. As a result, the full mask of the object can be decoded from image features updated by object slots. Extensive experiments on both real-world and synthetic benchmarks demonstrate the superiority of our proposed method, achieving state-of-the-art performance.
Our code will be released at \url{https://github.com/kfan21/EoRaS}.
\end{abstract}

\vspace{-0.1in}
\section{Introduction}
\label{sec:intro}

Deep learning has demonstrated remarkable success in various computer vision tasks. Nevertheless, neural networks are limited to learning visible patterns in the data, and are typically challenged in reasoning about the broader and unseen components. 
Currently, most researches in object detection and segmentation tasks concentrate on enhancing the visible part's performance, leaving few studies on inferring occluded information. 
Conversely, humans possess an innate ability to imagine and extrapolate, enabling us to easily complete an occluded part of an image based on prior knowledge. This critical capacity is instrumental in advanced deep learning models for real-world scenarios, such as medical diagnosis and autonomous driving. Thereby, the central issue addressed in this paper is the video amodal segmentation task, which aims to deduce an object's complete mask, whether it is partially obscured or not.

\begin{figure}
    \centering
    \includegraphics[width=0.95\linewidth]{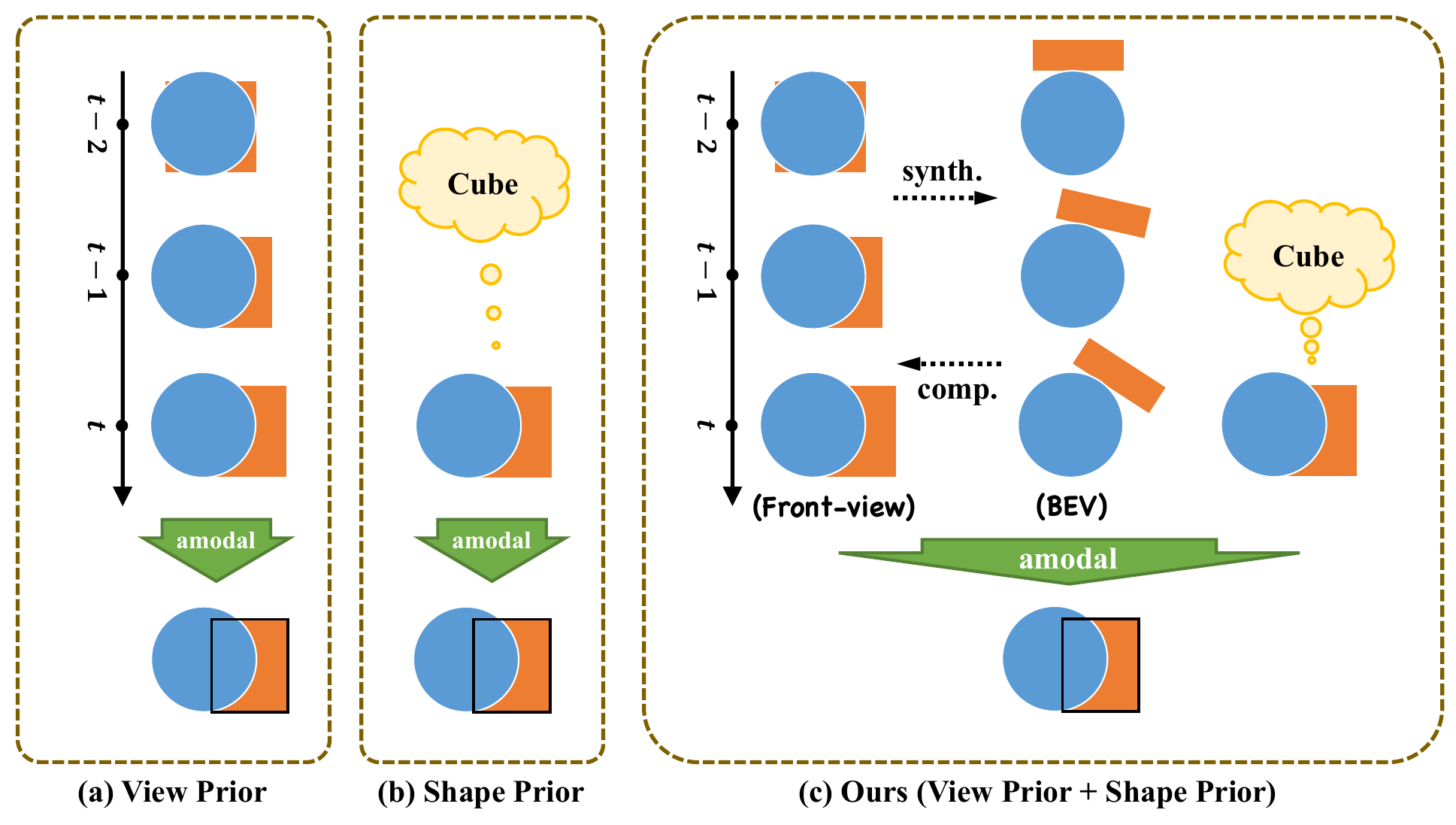}
    \vspace{-0.1in}
    \caption{Illustrations of the difference between view prior, shape prior, and our model. While SaVos~\cite{yao2022self} draws support from the optical flow to realize the view prior, image-level amodal segmentation algorithms typically just utilize the shape prior brought in by the supervision signals.
    Consequently, they are limited by camera motion and complicated object types, respectively. Unlike the previous methods, beyond the mergence of those two priors, \alg{} utilizes view prior by object-centric learning and further introduces the BEV space where obstruction doesn't exist, which enables our \alg{} to easily handle complex scenarios.
      \label{fig:motivation}
      }
  
    \vspace{-0.2in}
\end{figure}

Prior studies on image amodal segmentation~\cite{qi2019amodal, tran2022aisformer,  xiao2021amodal} 
are over-reliance on prior knowledge, which actually hampers the model's generalization abilities, resulting in limited improvements under complex circumstances. 
For video amodal, Yao et al.~\cite{yao2022self} proposed that the occluded part of the current frame may appear in other frames, and therefore, information from all frames should be collected to fill in the occluded regions of any specific frame. While this method achieves promising results under the \textit{self-supervised setting}, it fails when camera motion exists, as 2D warping is used to make connections within different frames, leading to distorted signals.

This paper aims to propose a better approach for video amodal segmentation by rethinking the importance of using supervised signals with object-centric representation. Such object-centric representations reflect the compositional nature of our world, and potentially facilitate supporting more complex tasks like reasoning about relations between objects.
While signals such as motion flow and shape priors have shown promising results, they are limited by moving cameras and complicated object types
respectively. In contrast, recent advances~\cite{deng2019restricted,hu2021istr,pan2020cross} in video object segmentation produce highly accurate object masks that are less sensitive to moving cameras, making them better suited as supervision signals. Surprisingly, such video object masks have not been fully exploited before.

To this end, we propose a novel approach that learns video amodal segmentation not only from observed object supervision signals in the current frame (\textit{shape prior}) but also from integrated information of object features under different views  (\textit{view prior}). Our motivation is clearly shown in Fig.~\ref{fig:motivation}. By using visual patterns of other views to explain away occluded object parts in the current frame~\cite{explain_away}, our approach gets rid of optical flow and eliminates the shortcomings of mere reliance on shape priors. Our model is highly effective, even in complex scenarios.

In this paper, we propose a novel supervised method for the video amodal segmentation task that leverages a multi-view fusion layer based temporal module and a Bird's-Eye View (BEV) feature translation network. Rather than relying on warping the amodal prediction into the next frame using optical flow or using shape priors alone, we enhance the current frame features by incorporating feature information from different viewpoints and leveraging the supervision signals simultaneously. Specifically, we first extract front-view features from the videos using FPN50~\cite{lin2017feature}. Then, we employ a translation network to transform these front-view features into bird's-eye view features, which bring in 3D information through the usage of the intrinsic matrix. In contrast to some related work~\cite{sharma2022seeing}  extracting object-centric 3D representation by object reconstruction, the acquisition of BEV feature is simpler, faster, and easier to train.
As each frame is equivalent to a unique view,
features from both different frames and the BEV space, which carry shape information about the occluded part, are further utilized. We repurpose the vanilla object-centric representations~\cite{locatello2020object} -- object slots to integrate those information, which is accomplished by our  novel multi-view fusion layer. Finally, we refine the front-view features using the updated object slots containing object information from multiple views and decode the full mask. Compared to previous methods~\cite{yao2022self}, our model can handle scenarios with 3D viewing angle changes or complex object shapes better by leveraging shape knowledge and integrating information across multiple views simultaneously.

To evaluate our method, we conduct extensive experiments on real-world and synthetic amodal benchmarks. The results demonstrate that our model achieves outstanding performance compared to comparable models and effectively demonstrates the efficacy of our architecture.

In summary, our main contributions are listed below.
(1) Our contribution lies in formulating the video amodal segmentation task using supervised signals for the first time. Our model efficiently learns the shape and view priors, enabling it to handle complex scenarios with ease.
(2) We propose a novel approach to learning object-centric representations through a multi-view fusion layer based temporal module equipped with a set of object slots, which achieves significant improvement in the correlation of information from different views.
(3) We introduce the novel concept of bird's-eye view features in our amodal task, which provides front-view features with 3D information, resulting in consistent benefits.
(4) By utilizing the bird's-eye view generator and multi-view fusion layer based temporal module, our algorithm achieves remarkable improvement on both real-world and synthetic amodal benchmarks, highlighting the novelty of our approach.

\section{Related Work}

\noindent\textbf{Amodal segmentation} is a more challenging task than instance segmentation because it requires predicting the full shape of occluded objects through the visible parts. While previous literature has focused on using shape priors effectively through multi-level coding~\cite{qi2019amodal}, variational autoencoder~\cite{jang2020learning}, shape prior memory codebook construction~\cite{xiao2021amodal}, 
mixing feature decoupling~\cite{ke2021deep} or Bayesian model~\cite{sun2022amodal}, 
relying solely on shape priors can lead to poor empirical performance due to distribution shifts between training data and real scenarios. To address this issue, \cite{yao2022self} leverages spatiotemporal consistency and dense object motion to explain away occlusion. Although their work has made progress in video amodal segmentation, optical flow can cause object deformation in the presence of camera motion. In contrast, our proposed architecture introduces a novel approach that does not require optical flow and utilizes bird's-eye view features to bring in 3D information that enhances the learning of front-view features.

\begin{figure*}
\centering
\includegraphics[width=0.9\linewidth]{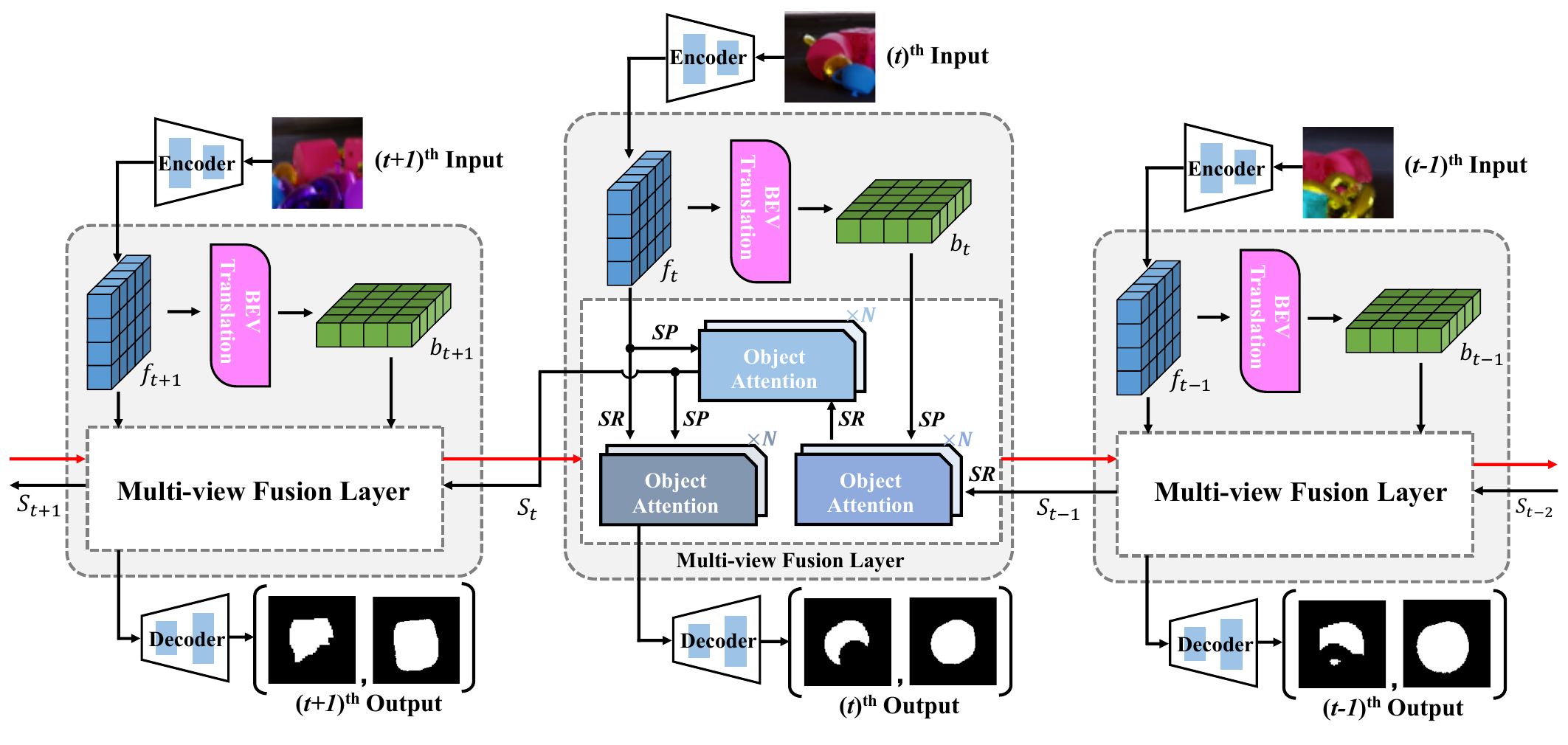}

   \vspace{-0.1in}
\caption{A schematic illustration of our method. The novelty of this architecture mainly lies in the BEV translation network and the multi-view fusion layer. The SP and SR represent the shape provider and receiver (see Section \ref{archi} for detail), respectively. \label{fig:method}}

   \vspace{-0.2in}
\end{figure*}

\noindent\textbf{Object-centric learning} aims at identifying all the objects from raw input for better understanding the complex scenes. Existing object-centric learning methods can be categorized into unsupervised and supervised methods. While unsupervised methods use image/scene reconstruction to extract object representations from images/scenes~\cite{burgess2019monet,locatello2020object,sajjadi2022object}, supervised methods represent each object as a query embedding and pay much attention to obtaining a great initialization~\cite{carion2020end,cheng2022masked,dong2021solq,fang2021instances,hu2021istr,yan2023universal}. Our \alg{} is more related to the supervised method in terms of constructing a set of learnable queries as an information container.

\noindent\textbf{BEV maps generation} requires to generate semantic maps in bird's-eye view space. Due to a lack of high-quality annotated data, most of the early work adopts weak supervision by utilizing stereo information~\cite{lu2019monocular,pan2020cross} or obtaining pseudo label~\cite{schulter2018learning}. Others directly translate semantic segmentation maps from image space into bird's-eye view space~\cite{deng2019restricted,samann2018efficient}. With the advent of large-scale annotated datasets, research on supervised methods has also made some progress. \cite{roddick2020predicting} and \cite{saha2022translating} respectively take advantage of dense transformer layer and 1D sequence-to-sequence translations to learn a map representation. \cite{berman2018lovasz} and \cite{li2022bevformer} instead blend features from multi-camera images to construct BEV map. In our \alg{}, the bird's-eye view feature is utilized to integrate 3D information into the front-view feature. To the best of our knowledge, it's the first attempt to incorporate the BEV translation module in the amodal segmentation task.

\section{Methodology}
This paper focuses on the video amodal segmentation task. Specifically, given a video sequence $\{I_{t}\}_{t=1}^{T}$ with $K$ objects, \alg{} aims to predict the full mask $\{M_{t}^{k}\}$ of each object in all frames, where $k$ is the index of objects. In our \alg{}, visible masks $\{V_{t}^{k}\}$ also serves as supervision  \textit{but will not be utilized at the test phase}. 

\subsection{Architecture}
\label{archi}

The overall architecture of \alg{} is shown in Figure~\ref{fig:method}. Our \alg{} is mainly comprised of four modules: (\textrm{i}) the feature encoding module which extracts the front-view feature $f_{t}^{k}$ from the input frames; (\textrm{ii}) the BEV translation network which converts the front-view features into bird's-eye view angle $b_{t}^{k}$ using the camera intrinsic matrix $K$ and neural network; (\textrm{iii}) the multi-view fusion layer based temporal module which utilizes the object slots updated through the forward and backward streams to integrate the feature information from different views and fulfill the completion of each front-view feature;
and (\textrm{iv}) the deconvolution network that estimates the full masks and visible masks of the current frame simultaneously.

\noindent\textbf{Feature Encoding Module} In this module, FPN50~\cite{lin2017feature} pretrained on ImageNet~\cite{deng2009imagenet} is used to extract features from the input frames. These features are obtained from a frontal perspective and capture a lot of information but will fail to make inferences about the missing parts of the objects. 

\begin{equation}
f_{t}^{k} = FPN(I_{t}^{k}) 
\end{equation}

\begin{figure*}
\centering
    \begin{subfigure}{0.48\textwidth}
    \centering
    \includegraphics[width=1\textwidth,height=0.48\textwidth]{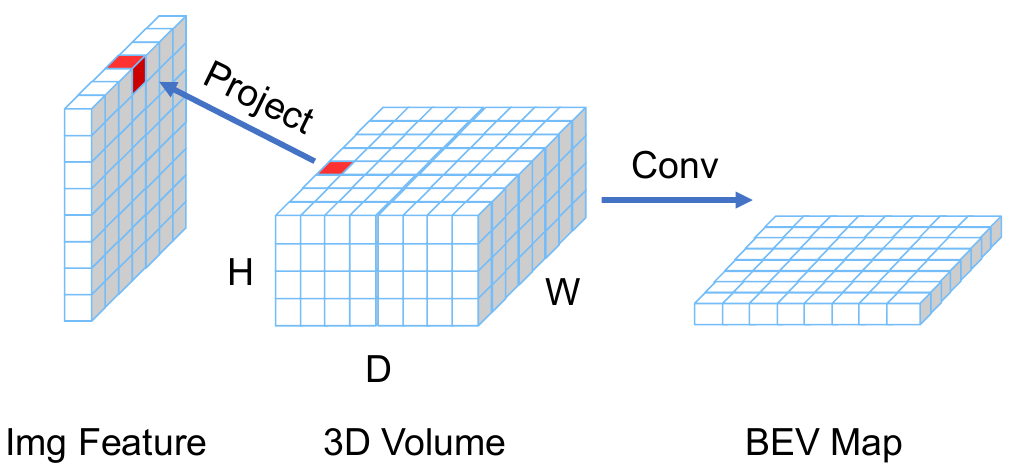}
    \caption{BEV Translation Network}
    \label{fig:bev}
    \end{subfigure}
    \begin{subfigure}{0.48\textwidth}
    \centering    \includegraphics[height=0.48\textwidth]{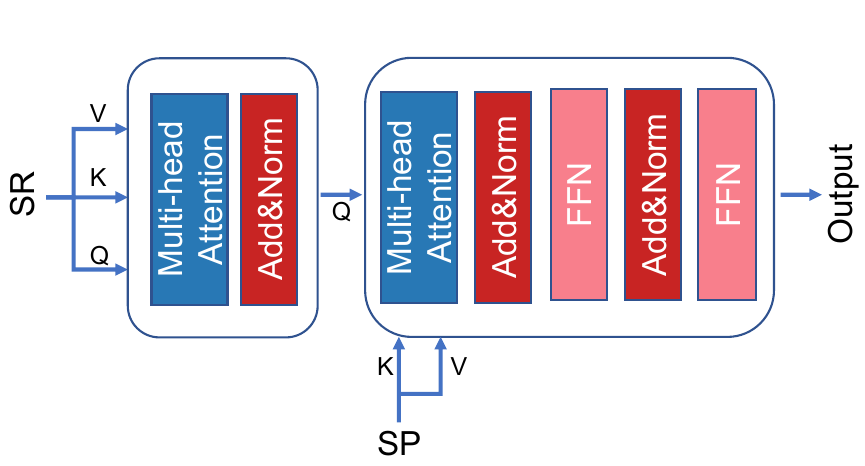}
    \caption{Object Attention Layer}
    \label{fig:attention}
    \end{subfigure}
\vspace{-0.1in}
\caption{(a) A three-dimensional cuboid is built and rasterized in the camera coordinate. For each voxel, we use the intrinsic matrix to obtain its coordinates in the plane system, and use bilinear interpolation on the front-view features to obtain its feature. Then, a convolutional network is used to obtain the bev map. (b) Our object attention layer in the multi-view fusion layer is stacked by self-attention, cross-attention, and feedforward network. This layer is designed for shape information fusion and takes two variables as input. We nominate the variable to be updated as shape receiver (SR) and another one as shape provider (SP).
}
\vspace{-0.2in}
\end{figure*}

\noindent\textbf{BEV Translation Network} The features from the bird's-eye view (BEV) are widely used and work well in autonomous driving research. Recall that features from different perspectives are likely to contain the missing part information and contribute to the full mask deduction of the current frame. As obstruct doesn't exist in BEV space unless objects are stacked on top of each other, it is reasonable to  introduce BEV as a special perspective to promote the completion of the front-view feature. Consider a horizontally placed camera, for each frame,
a 3D volume feature $V_{3D}$ is constructed in the camera coordinate system. As the BEV feature generation just involves the current frame, we omit the subscript of time $t$ and the object index $k$ for simplicity.

Denote the camera intrinsic matrices as $K$, we first focus on a single point $(x,y,z)$ in the camera coordinate space. By utilizing intrinsic matrices, this point can be easily projected into the image/feature plane and we denote its coordinate as $(u, v)$:
\begin{equation}
\left(\begin{array}{c}
\lambda u \\
\lambda v \\
\lambda
\end{array}\right)=K\left(\begin{array}{c}
x \\
y \\
z
\end{array}\right)
\end{equation}
We use bilinear interpolate to obtain the feature at $(u, v)$ from the corresponding front-view feature $f$. The obtained value at $(u, v)$ will act as the volume feature at $(x, y, z)$.

As shown in Figure~\ref{fig:bev}, the 3D volume in the \textit{camera coordinate} will be rasterized into a group of points $p_{ijk}=(x_i, y_j, z_k)$, where $1\le i\le m, 1\le j\le n, 1\le k\le h$ and $x_i,y_j,z_k$ are three predefined 1D grid. $x, y, z$ represent the direction of width, depth, and height, respectively. The value of $V_{3D}$ will be gotten by simply repeating the above process for each point. Further, by stacking the feature of the volume obtained from different channels together, we will get $V_{3D}\in\mathbb{R}^{c\times m\times n\times h}$. Since our goal is to acquire BEV features, $V_{3D}$ is rearranged to $\mathbb{R}^{ch\times m\times n}$ and sent to a lightweight CNN for compression along  height dimension:
\begin{equation}
b_{k}^{t} = \mathrm{CNN}(V_{3D}.reshape(ch,m,n))
\end{equation}

\noindent\textbf{Multi-view Fusion Layer based Temporal Encoder} 
As the occluded part of a specific view may potentially appear in other frames, we can make full use of the information in each frame (equivalent to different perspectives) to refine 
 the completion of the object shape. Specially, inspired by DETR~\cite{carion2020end} and Slot Attention~\cite{locatello2020object}, we would like to generate an object-centric feature utilizing both front-view and BEV representations. 

A direct method is to follow \cite{locatello2020object}, which uses ConvGRU to aggregate temporal information. However, the cost of nested recurrent slot computation to gather the object information from each frame is expensive when processing videos. Here, we propose a more efficient attention-based encoder architecture named Multi-view Fusion Layer.  
Generally, in such a layer, three $N$-layer object attentions which is a non-recurrent variant of slot attention are
carefully designed and closely connected. And features from different views and object slots serve as the inputs.

In particular, as shown in Figure~\ref{fig:attention}, each object attention layer ($\operatorname{ObjAttention}(SP, SR)$) is stacked by self-attention, cross-attention, and feed-forward networks and serves as information fusion network. The variable absorbing the missing shape information during the fusion process is named shape receiver (SR), and another one is dubbed shape provider (SP) as it offers extra shape patches. The total forward process in $\operatorname{ObjAttention}$ is formulated as,

\begin{equation}
    \hat{SR} = SR + \operatorname{Attention}(SR, SR, SR)
\end{equation}
\begin{equation}
    \widetilde{SR} = \hat{SR} + \operatorname{Attention}(SP, \hat{SR}, SP)
\end{equation}
\begin{equation}
    output =  \operatorname{MLP}(\widetilde{SR})
\end{equation}
where $\operatorname{Attention}(K,Q,V), \operatorname{MLP}(\cdot)$ denotes multi-head attention module and two-layer feedforward network, respectively. 
And we omit all normalization layers. $\operatorname{ObjAttention}$ first enhances the SR representation by renewing information contained in itself, then extracts fresh properties from the SP. 

On the other hand, similar to \cite{carion2020end}, a set of object slots $S_{0} \in R^{n_{s} \times d}$ is initialized before the videos enter. $n_{s}$ denotes the number of slots and $d$ is the feature dimension. In our model, $S_{0}$ is set to be learnable and serves as a container that gathers shape information from various views. 

With the above preparations, we now go to the detail of our multi-view fusion layer. For each frame, we take advantage of the $S_{t-1}$ from the last frame which includes object shape information from previous frames, and provide it with the fresh characters from the front-view $f_{t}^{k}$ and BEV feature $b_{t}^{k}$ under current perspective at first:
\begin{equation}
S_{t}^{\prime} = \operatorname{ObjAttention}(SR=S_{t-1}, SP=b_{t}^{k})
\end{equation}
\begin{equation}
S_{t} = \operatorname{ObjAttention}(SR=S_{t}^{\prime}, SP=f_{t}^{k})
\end{equation}
Then, the updated slots will provide clues about the occluded part and help complete the front-view features of the current frame by setting the front-view features as SR in the object attention layer. Thus, we inversely enhance the front-view feature using the object slots by:
\begin{equation}
\hat{f}_{t}^{k} = \operatorname{ObjAttention}(SR=f_{t}^{k}, SP=S_{t})
\end{equation}

\noindent\textbf{Deconvolution Network} The deconvolution network (DeConv) is served as the mask predictor and takes the updated front-view features as input since it shares the same perspective with the full mask to be predicted. In our experiments, we just construct several de-convolutional layers for this module. 
\begin{equation}
\hat{M}_{t}^{k}, \hat{V}_{t}^{k} = \mathrm{DeConv}(\hat{f}_{t}^{k})
\end{equation}
where $\hat{M}_{t}^{k}$ and $\hat{V}_{t}^{k}$ are the full and visible mask predictions of the current frame, respectively. 

\noindent\textbf{Bi-directional Prediction} Cold start problem exists under the above framework since the first few frames may not be informative enough. Thus, backward prediction is added to solve this problem.
We simply concatenate the forward and backward features, and send them to the final deconvolution network.

\subsection{Loss Function for \alg{}} 
\label{sec:loss_function}
Our \alg{} is designed as an end-to-end framework and trained with the focal loss ($\mathrm{Focal}()$) using both full mask and visible mask as supervision signals. Note that the discard of the visible mask loss will not heavily damage the model performance, as shown in Tab.~\ref{tab:ablation_lambda}. 
The overall loss function is
\begin{equation}
\mathcal{L}_{full} = \sum_{t=1}^{T} \sum_{k=1}^{K} \mathrm{Focal}(\hat{M}_{t}^{k}, {M}_{t}^{k})
\end{equation}
\begin{equation}
\mathcal{L}_{vis} = \sum_{t=1}^{T} \sum_{k=1}^{K} \mathrm{Focal}(\hat{V}_{t}^{k}, {V}_{t}^{k})
\end{equation}

\begin{equation}
\mathcal{L} = \mathcal{L}_{full} + \lambda \cdot \mathcal{L}_{vis} .
\end{equation}

\section{Experiments}

\begin{figure}
  \centering
  \includegraphics[width=0.43\textwidth]{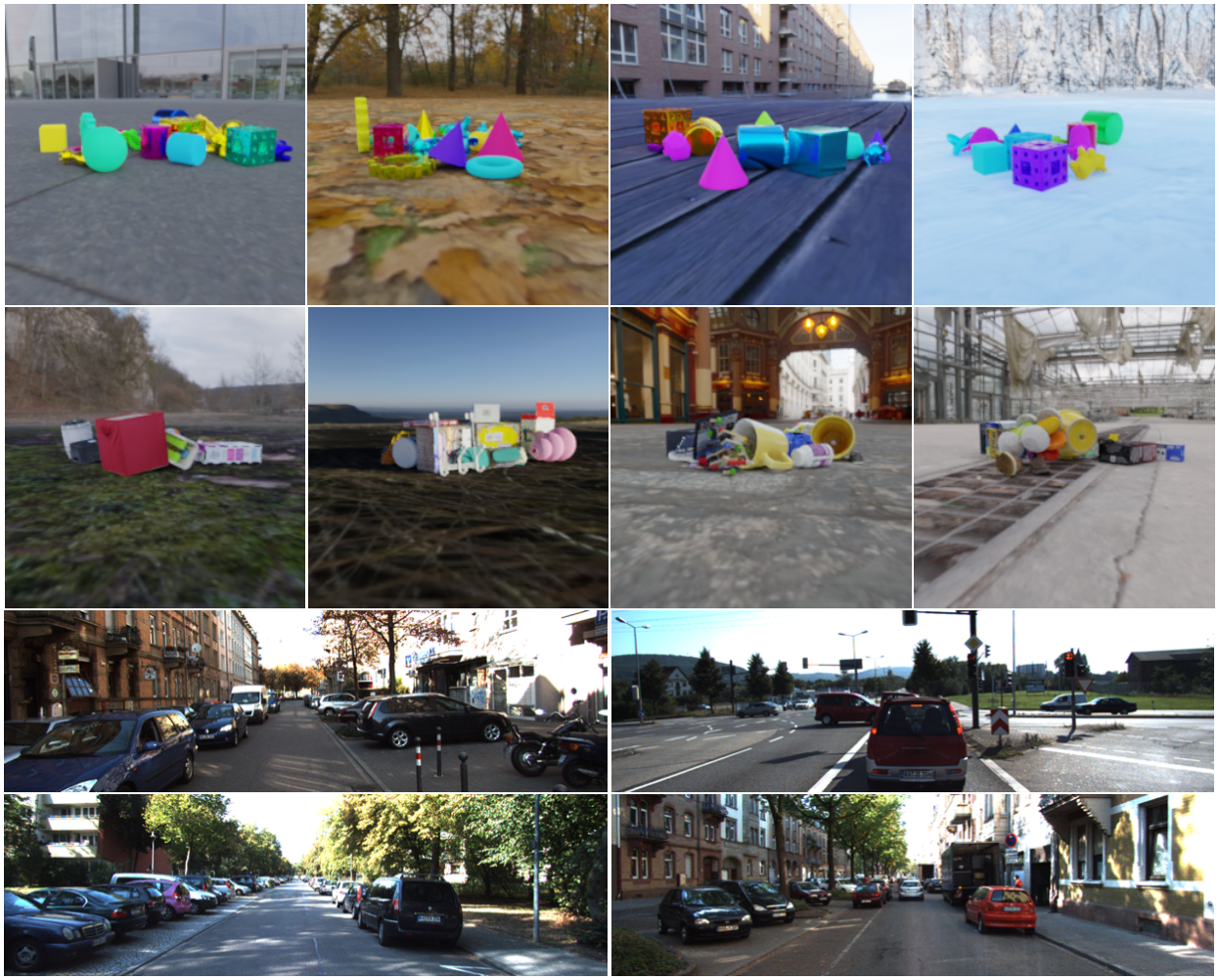}
     \vspace{-0.1in}
  \caption{Visualization of datasets. The first and second rows show images from the Movi-B and Movi-D, respectively. The remaining four images belong to the KITTI.   \label{fig:dataset}}

   \vspace{-0.2in}
\end{figure}

To fully evaluate our model, we conduct extensive experiments on both real-world and synthetic amodal segmentation benchmarks, including Movi-B, Movi-D, and KITTI datasets, with the visualization in Fig.~\ref{fig:dataset}. 

\noindent\textbf{Movi Dataset}~\cite{greff2021kubric} is a \textit{synthetic} dataset consisting of random scenes and objects created by Kubric~\cite{greff2021kubric}. In our experiments, we consider two datasets (Movi-B and Movi-D) with different objects and different levels of occlusions. \textit{We extract the amodal information during generation of the two datasets}. The objects in Movi-B and Movi-D are from the CLEVR~\cite{johnson2017clevr}, which consists of 11 relatively regular object shapes, and Google Scanned Objects~\cite{downs2022google}, which contains 1030 realistic objects, respectively. Both datasets use the background from Poly Haven. To create situations with serious occlusion, all objects are set to be static and stacked closely together. Videos are created by setting the camera to rotate around the objects. Overall, compared with Movi-B, Movi-D has a more complex object shape and lower camera viewing angle with more serious occlusion.

\begin{figure*}[htbp]
  \centering
  \includegraphics[width=0.95\textwidth]{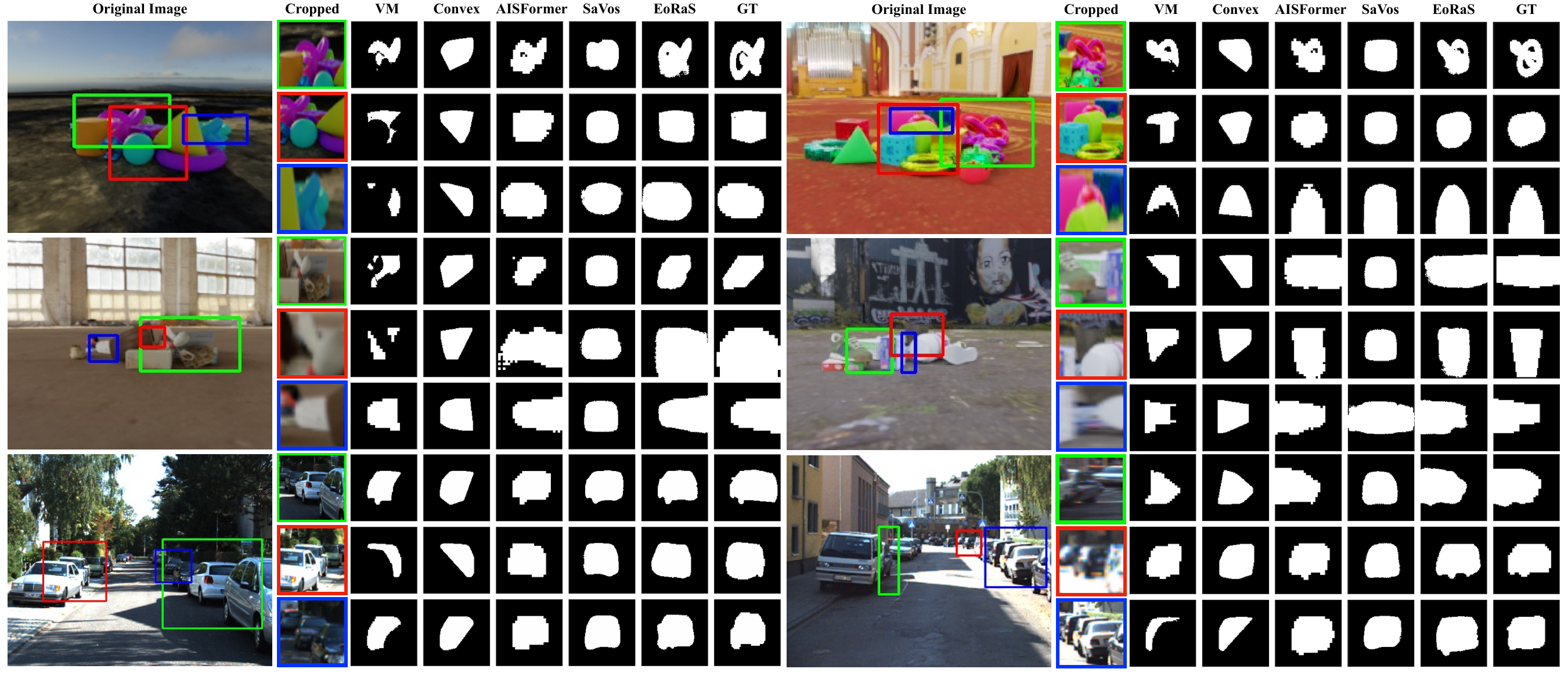}
      \vspace{-0.1in}
  \caption{Qualitative comparison between our \alg{} and competitors. From top to down, the three rows are from Movi-B, Movi-D, and KITTI, respectively.  \label{fig:visual_mask} }
 
      \vspace{-0.15in}
\end{figure*}

\noindent\textbf{KITTI Dataset}~\cite{Geiger2013IJRR} is currently the largest real-world autonomous driving evaluation dataset. It has been widely used in many vision tasks, such as object detection and optical flow prediction. ~\cite{qi2019amodal} annotated some images in KITTI with amodal information and \cite{yao2022self} matched these images to its original video frame. Note that since these videos are not sufficiently annotated, it is a weakly supervised scenario. For a fair comparison, we follow the same data split in~\cite{yao2022self}. The visible masks and object tracks are extracted by PointTrack \cite{xu2020Segment}. It is noteworthy that only the car category is annotated in this dataset.

\begin{table} \small 
      \centering
        \begin{tabular}{c | c | c c}
        \midrule
            \multirow{2}{*}{\textsc{DATASET}}
           & \multirow{2}{*}{\textsc{METHODS}} & \multicolumn{2}{c}{\textsc{Metrics}} \\
            \cline{3-4}
           & & mIoU$_{full}$ & mIoU$_{occ}$\\
         \midrule
          \multirow{7}{*}{Movi-B} & VM & 59.19 & - \\
          & Convex & 64.21 & 18.42\\
          & PCNET & 65.79 & 24.02\\
          & AISFormer & 77.34 & 43.53\\
          & SaVos-sup. & 70.72 & 33.61\\
          & BiLSTM & 77.93 & 46.21\\
          \cline{2-4}
          & \alg{} \textit{(Ours)} & \textbf{79.22} & \textbf{47.89}\\
          \cline{1-4}
          \multirow{7}{*}{Movi-D} & VM & 56.92 & - \\
          & Convex & 60.18 & 16.48\\
          & PCNET & 64.35 & 27.31\\
          & AISFormer & 67.72 & 33.65\\
          & SaVos-sup. & 60.61 & 22.64\\
          & BiLSTM & 68.43 & 36.00\\
          \cline{2-4}
          & \alg{} \textit{(Ours)} & \textbf{69.44} & \textbf{36.96}\\
          \cline{1-4}
            \multirow{7}{*}{KITTI} & VM & 74.75 & - \\
          & Convex & 78.62 & 8.29\\
          & PCNET & 81.58 & 17.90\\
          & AISFormer & 86.42 & 51.04\\
          & SaVos-sup. & 83.09 & 37.33\\
          & BiLSTM & 86.68 & 49.95\\
          \cline{2-4}
          & \alg{} \textit{(Ours)} & \textbf{87.07} & \textbf{52.00}\\

           \midrule
        \end{tabular}
     \vspace{-0.1in}
    \caption{The performance of \alg{} on real-world and synthetic video amodal benchmarks.   \label{tab:movi+kitti}} 
 
         \vspace{-0.12in}
\end{table}

\subsection{Competitors and Settings}
\noindent\textbf{Competitors} We compare our method with the following related methods:
(1) \textbf{VM (Visible Mask)}, directly use the ground truth visible mask as amodal prediction;  (2) \textbf{Convex},  take the convex hull of the visible mask as the amodal mask;
(3) \textbf{PCNET}\cite{zhan2020self},  a self-supervised image-level amodal completion method by in turn recovering occlusion ordering and completing amodal masks and content;
(4) \textbf{AISFormer}~\cite{tran2022aisformer}, an image-level amodal segmentation model equipped with a transformer-based mask head and achieves the new state-of-the-art recently;
(5) \textbf{Savos}~\cite{yao2022self}, a recent state-of-the-art method in the field of self-supervised video amodal segmentation and is modified to supervised version by removing the 2D warping and bringing the supervised signal for fair comparison (We also did additional experiments involving warping operation, but the experiment results are quite inferior);
(6) \textbf{BiLSTM}~\cite{hochreiter1997long}, a variant of our proposed method for which we keep the same FPN50 backbone but utilize BiLSTM to aggregate temporal information across frames.

\noindent\textbf{Implementations} Results on all datasets are reported in terms of mIOU metrics for both full mask and occluded regions. Since most amodal segmentation algorithms use the visible mask or the bounding boxes of the visible part as model input, the estimation results of the visible area may be more confident, and the mIOU of the occluded area can better reflect the model performance. 
On all datasets, the mIOU metric of the occluded part is only computed on those partially occluded objects. We use AdamW as optimizer with batch size 4 for 50 epochs. The learning rate is set to $1e-5$ on Movi datasets and $1e-4$ on the KITTI dataset. Exponential learning rate decay is used where the decay rate is 0.95. The weight decay is $5e-4$. And the $\gamma$ in focal loss is set to 2. We set $\lambda=1$, $n_{s}=8$, $N=2$ and train our model on four Tesla T4 GPUs using PyTorch.

\begin{table*} \small 
    \begin{subtable}{.5\linewidth}
      \centering
        \setlength{\tabcolsep}{1mm}{
        \begin{tabular}{c | c c c | c c}
        \midrule
        
          \multirow{2}{*}{\textsc{No.}} & \multicolumn{3}{c|}{\textsc{Designs}} & \multicolumn{2}{c}{\textsc{Metrics}} \\
          \cline{2-6}
           & Temporal & Bi-direction & BEV & mIoU$_{full}$ & mIoU$_{occ}$ \\
           
          \midrule
           1 & $\times$ & $\times$ & $\times$ & 76.93 & 44.55 \\
           2 & \checkmark & \checkmark & $\times$ & 78.66 & 46.83 \\
           3 & \checkmark & $\times$ & \checkmark & 78.42 & 46.51 \\
           4 & \checkmark & \checkmark & \checkmark & \textbf{79.22} & \textbf{47.89} \\
          \midrule
        \end{tabular}
        }
    \end{subtable}%
    \begin{subtable}{.5\linewidth}
      \centering
        \setlength{\tabcolsep}{1mm}{
        \begin{tabular}{c | c c c | c c}
        \midrule
        \multirow{2}{*}{\textsc{No.}} & \multicolumn{3}{c|}{\textsc{Designs}} & \multicolumn{2}{c}{\textsc{Metrics}} \\
          \cline{2-6}
           & Temporal & Bi-direction & BEV & mIoU$_{full}$ & mIoU$_{occ}$ \\
          \midrule
          5 & $\times$ & $\times$ & $\times$ & 66.70 & 33.42 \\
          6 & \checkmark & \checkmark & $\times$ & 69.08 & 36.39 \\
          7 & \checkmark & $\times$ & \checkmark & 68.56 & 35.54 \\
          8 & \checkmark & \checkmark & \checkmark & \textbf{69.44} & \textbf{36.96} \\
          \midrule
        \end{tabular}
        }
    \end{subtable} 
     \vspace{-0.1in}
    \caption{Ablation study of our temporal and bev modules on Movi-B (left) and Movi-D (right) dataset.   \label{tab:ablation_movi}} 
 
         \vspace{-0.15in}
\end{table*}

\subsection{Results on Movi Datasets}

As shown in Table \ref{tab:movi+kitti}, compared with supervised SaVos, our \alg{} achieves extremely significant performance improvements on both Movi datasets. In particular, by applying our algorithm, the prediction of the full mask of the objects in the two datasets is improved by 8.50\% and 8.83\%, respectively. The improvements are more remarkable in the prediction of occluded parts. For the performance on the occluded part, our \alg{} achieves 14.28\% improvement on the Movi-B over the baseline SaVos, and surprisingly improves by 14.32\% on the Movi-D. Moreover, the performance of \alg{} also exceeds the recent state-of-the-art image-level algorithm AISFormer by a clear margin on both datasets. And it's noteworthy that \alg{} outperforms the combination of FPN50 and BiLSTM/Transformer by at least 1\% in plenty of experiments, showing the effectiveness of introducing the BEV module. Additionally, despite the usage of ground truth visible mask in Convex and PCNET, \alg{} still exhibits amazing power. We also present the qualitative results in Figure \ref{fig:visual_mask}. Obviously, the full masks deduced by \alg{} are the closest to the original object shape among all the competitors. Above all, \alg{} is more suitable for solving the video amodal segmentation task, and leads to the new state-of-the-art.

\begin{figure}
\centering
    \includegraphics[width=0.45\textwidth]{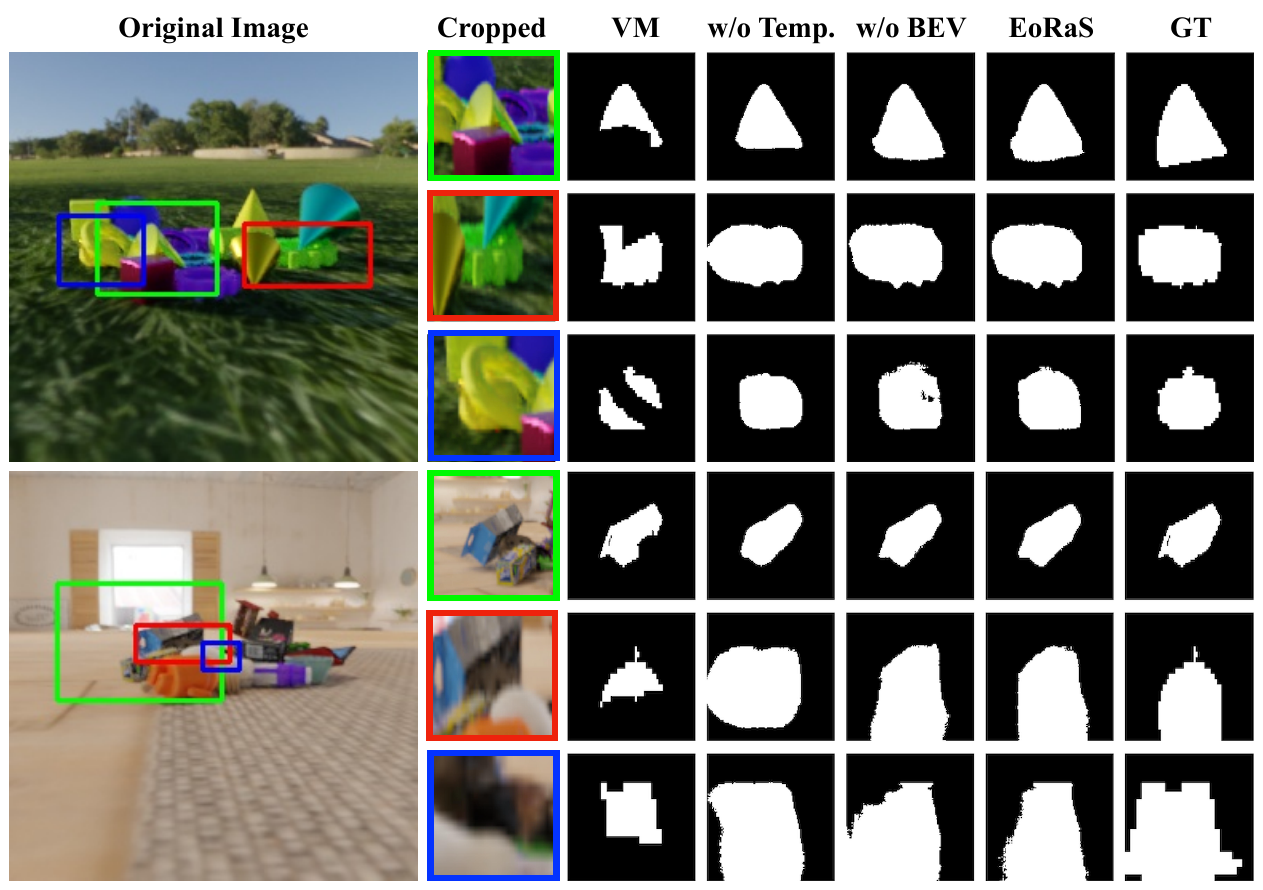}
    \vspace{-0.1in}
    \caption{Visualizations derived from models with/without proposed temporal or BEV module. Clearly, the complete model deduces the best full mask, and both the temporal and BEV module can bring consistent benefits.}
    \label{fig:visual_abla}
    \vspace{-0.2in}
\end{figure}

\subsection{Results on KITTI Dataset}

The experiment results on the KITTI dataset are shown in Table \ref{tab:movi+kitti}. For objects in real scenes, our \alg{} can still exceed all the current state-of-the-art methods. Compared with the image-level baseline, we achieve 0.65\% and 0.96\% improvement for the full and occluded mask prediction, respectively. For the supervised SaVos, \alg{} achieves enormous promotion, $\sim$4\% on the full shape and $\sim$15\% on the missing part. Furthermore, other video-level baselines consistently underperform our \alg{} by $\sim$2\% on the deduction of the occluded part. Qualitative comparison in Figure \ref{fig:visual_mask} clearly exhibits the great precision of \alg{}. The above evidence is sufficient enough to prove the effectiveness of \alg{} under weakly supervised settings.

\begin{table} \small
    \begin{center}
    \setlength{\tabcolsep}{3mm}{
    \begin{tabular}{c|c| c c}
    \midrule
      \textsc{Dataset} & \# Slots & mIoU$_{full}$ & mIoU$_{occ}$ \\
        \midrule
       \multirow{6}{*}{Movi-B}
       & 8   & 79.22 & 47.89 \\
       & 16  & 79.22 & 47.79 \\
       & 32  & 79.29 & 47.88 \\
       & 64  & 79.22 & 47.78 \\
       & 128 & 79.19 & 47.73 \\       
       & 256 & 79.20 & 47.75 \\  
      \hline
       \multirow{6}{*}{Movi-D}
       & 8   & 69.44 & 36.96 \\
       & 16  & 69.42 & 36.92 \\
       & 32  & 69.50 & 37.27 \\
       & 64  & 69.38 & 37.01 \\
       & 128 & 69.47 & 37.02 \\
       & 256 & 69.45 & 37.06 \\       
      \midrule
    \end{tabular}}
    \end{center}
    \vspace{-0.25in}
    \caption{Sensitivity Analysis of Slot Number. Despite the diverse settings of slot number, the performance of \alg{} just changes slightly, demonstrating the robustness against hyper-parameter $n_{s}$.     \label{tab:ablation_slot_num}}
     \vspace{-0.25in}
\end{table}

\section{Further Analysis}

\noindent\textbf{Effectiveness of Temporal and BEV Modules} As shown in Table \ref{tab:ablation_movi}, on Movi-B dataset, our temporal module brings about $\sim$2.3\% performance improvement in occluded part prediction. After plugging in the BEV module, the occluded mIOU is further improved by 1.06\%. Additionally, Bi-direction prediction also plays an important role in our model as it brings 1.38\% performance improvement for the missing part deduction. On the Movi-D dataset, the improvements brought in by those modules are also significant, as presented in the right table. Some visualizations derived from different architectures are presented in Figure \ref{fig:visual_abla}. It's clear that both temporal and BEV modules are capable of improving the smoothness and shape similarity of full masks. These experiments fully prove the effectiveness of the modules proposed in this paper, and also verify the correctness of our hypothesis that feature information from different perspectives can benefit the completion of object shape in any specific frame/view.

\noindent\textbf{Sensitivity Analysis of Slot Number} To analyze the sensitivity to the choice of $n_{s}$, we conduct experiments by widely tuning the slot number. The results are presented in Table \ref{tab:ablation_slot_num} and indicate that the number of slots has almost no impact on the performance of our model. This phenomenon demonstrates the robustness of our model against the diverse choices of slot numbers.

\begin{table}\small
  \begin{center}
  \setlength{\tabcolsep}{3mm}{
  \begin{tabular}{c| c| c c}
  \midrule
    \textsc{Dataset} & $\lambda$ & mIoU$_{full}$ & mIoU$_{occ}$ \\
      \midrule
     \multirow{5}{*}{Movi-B}
     & 0.0  & 78.93 & 47.55 \\
     & 0.25 & 79.14 & 47.80 \\
     & 0.5  & 79.21 & 47.90 \\
     & 0.75 & 79.20 & 47.85 \\       
     & 1.0  & 79.22 & 47.89 \\
    \hline
     \multirow{5}{*}{Movi-D}
     & 0.0  & 68.68 & 36.39 \\
     & 0.25 & 69.26 & 37.06 \\
     & 0.5  & 69.42 & 36.99 \\
     & 0.75 & 69.38 & 36.96 \\    
     & 1.0  & 69.44 & 36.96 \\
    \midrule
  \end{tabular}}
  \end{center}
  \vspace{-0.25in}
  \caption{Performance of \alg{} under different $\lambda$.   \label{tab:ablation_lambda}}
     \vspace{-0.1in}
\end{table}

\begin{table}\small
  \begin{center}
  \setlength{\tabcolsep}{3mm}{
  \begin{tabular}{l| c| c c}
  \midrule  
    \multirow{2}{*}{\textsc{Methods}} & \multirow{2}{*}{\textsc{Target}} & \multicolumn{2}{c}{\textsc{Metrics}} \\
    \cline{3-4}
    & & mIoU$_{full}$ & mIoU$_{occ}$ \\
      \midrule
     \multirow{2}{*}{AISFormer \cite{tran2022aisformer}}
     & Movi-D  & 62.94 & 28.65 \\
     & KITTI & 71.36 & 29.84 \\
    \hline
     \multirow{2}{*}{SaVos-Sup. \cite{yao2022self}}
     & Movi-D  & 57.19 & 25.85 \\
     & KITTI & 65.49 & 21.82 \\
    \hline
    \multirow{2}{*}{\alg{} (\textit{Ours})}
    & Movi-D  & \textbf{63.98} & \textbf{31.22} \\
    & KITTI & \textbf{71.73} & \textbf{31.35} \\
   \midrule
  \end{tabular}}
  \end{center}
  \vspace{-0.25in}
  \caption{Open set segmentation on Movi-D and KITTI datasets. We use \alg{} pretrained on the Movi-B dataset and conduct transfer learning experiments without finetuning. Our \alg{} achieves the highest performance, indicating its great generalization ability.  \label{tab:transfer_study}}
     \vspace{-0.1in}
\end{table}

\begin{table}[t] \small
      \centering
        \setlength{\tabcolsep}{3.5mm}{
        \begin{tabular}{c | c  | c c}
          \midrule
            
            \multirow{2}{*}{\textsc{Dataset}} & \multirow{2}{*}{\textsc{Methods}} & \multicolumn{2}{c}{\textsc{Metrics}} \\
            \cline{3-4}
            & & mIoU$_{full}$ & mIoU$_{occ}$ \\
            \midrule
             \multirow{4}{*}{Movi-B} & \alg{} & 79.22 & 47.89 \\
             
             & +PP$^{*}$ & 79.38 & 47.66 \\
             & +PP & 81.20 & 47.89 \\
             & +SG & \textbf{81.76} & \textbf{49.39} \\
            \hline
            \multirow{4}{*}{Movi-D} & \alg{} & 69.44 & 36.96 \\
             & +PP$^{*}$ & 69.95 & 36.81 \\
             & +PP & 72.76 & 36.96 \\
             & +SG & \textbf{74.10} & \textbf{38.33} \\
            \midrule
          \end{tabular}}
    \vspace{-0.1in}
    \caption{The performance of \alg{} while using GTVM at test phase on Movi dataset. PP$^{*}$ and PP means the predicted and ground truth visible mask are used in post-process, respectively. And SG represents the model trained with the concatenation of images and visible masks.  \label{tab:ablation_process_movi}} 
   
       \vspace{-0.25in}
\end{table}

\noindent\textbf{Different choices of $\lambda$} 
We conduct experiments to analyze the effect of $\lambda$ on the performance of our model, and the results are presented in Table \ref{tab:ablation_lambda}. First of all, the utilization of visible masks in supervision signals will benefit the model training as also shown in previous amodal segmentation algorithms. But the way that \alg{} differs lies in the insensitivity to the choice of $\lambda$ once the visible mask is added, which demonstrates the superiority of \alg{}.

\noindent\textbf{Open Set Segmentation} To evaluate the capacity of out-of-distribution generalization, we conduct open set segmentation experiments on Movi-D and KITTI datasets. Models are pretrained on the relatively simple Movi-B dataset. As presented in Table~\ref{tab:transfer_study}, \alg{} achieves the best accuracy among all competitors. Concretely, compared with supervised SaVos, \alg{} outperforms by at least 6\%, showing strong dominance. Again, the image-level SOTA algorithm underperforms \alg{} by $\sim$2\% on the occluded part deduction, indicating that the integration of information from different views indeed benefits the generalization ability.

\noindent\textbf{Test-time Assistance by Ground Truth Visible Mask (GTVM)} The same as SaVos and PCNET, we explore the utilization of GTVM at the test phase. On the one hand, the post-processing (PP), including taking the intersection of the predicted full mask and GTVM, is feasible. On the other hand, containing partial shape information, GTVM may be capable of serving as a shape guidance (SG) for mask completion. To this end, we simply train our model with the concatenation of images and visible masks. The experimental results are presented in Table \ref{tab:ablation_process_movi}. Overall, the introduction of GTVM brings in huge benefits, which is inline with~\cite{yao2022self,zhan2020self}. Despite the usage of GTVM in those algorithms, our \alg{} still outperforms them by a large margin (see Table \ref{tab:movi+kitti}), suggesting the powerful function.

\section{Conclusion}
In this paper, we proposed a brand-new pipeline named \alg{} to cope with the video amodal segmentation task. Based on the assumption that both the supervision signals (shape prior) and the features from different perspectives (view prior) will benefit the deduction of the full mask under any specific view, the multi-view fusion layer based temporal encoder and BEV translation network are designed to integrate 3D information and front-view shape patches from different frames respectively in an object-centric pattern. Utilizing those modules, our \alg{} eliminates the optical flow usage and the over-reliance on shape priors, achieving high efficiency even in complex scenarios. We conduct experiments on both real-world and synthetic video amodal benchmarks, including Movi-B, Movi-D, and KITTI datasets. The empirical results demonstrate that our \alg{} achieves the new state-of-the-art performance.

\noindent {\small \textbf{Acknowledgements.}  This work is supported by China Postdoctoral Science Foundation (2022M710746). Yanwei Fu is with the School of Data Science,  Shanghai Key Lab of Intelligent Information Processing, Fudan University, and Fudan ISTBI—ZJNU Algorithm Centre for Brain-inspired Intelligence, Zhejiang Normal University, Jinhua, China.}

{\small
\bibliographystyle{ieee_fullname}
\bibliography{camera_ready}
}

\clearpage
\appendix
\onecolumn

\section{Qualitative comparison between \alg{} and competitors}

In the supplementary part, we show some qualitative comparisons of our model and competitors. 

\subsection{Qualitative comparison in Movi-B and Movi-D}

\begin{figure*}[!h]
\centering
\includegraphics[width=0.98\linewidth]{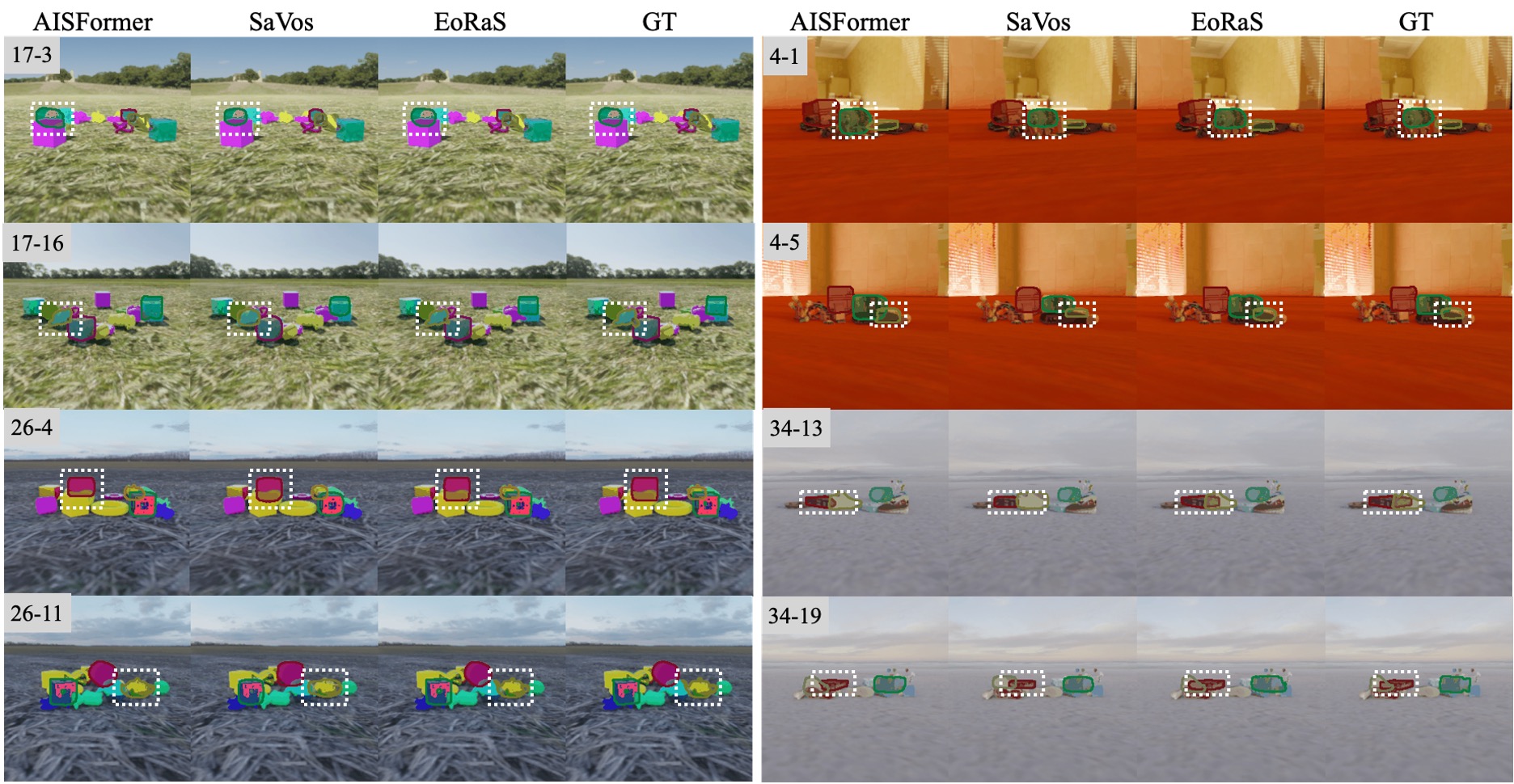}
   \vspace{-0.1in}
\caption{
Qualitative comparison between our \alg{} and competitors in the Movi-B and Movi-D datasets. The images in the left column are from Movi-B, and those in the right column are from Movi-D. The numbers in each upper-left corner indicate where these images come from. For example, 17-3 indicates this image is from the $3^{rd}$ frame of the $17^{th}$ video. For convenience, we also put frames on those objects with the largest predicted mask difference.
}
\label{fig:movi}
\vspace{-0.2in}
\end{figure*}

Figure \ref{fig:movi} provides a comprehensive qualitative comparison between our \alg{} and competitors across two datasets, Movi-B and Movi-D. The left column displays images from Movi-B, and the right column displays images from Movi-D, with the numbers in the upper-left corner indicating the source frame of each image. For example, 17-3 indicates this image is from the $3^{rd}$ frame of the $17^{th}$ video. Notably, we also highlight the objects with the largest predicted mask difference by framing them for ease of comparison.

When analyzing the images from Movi-B, our model outperforms competitors in many cases. For example, in the first image (17-3), our prediction for the green cylinder is superior to those of our competitors. Specifically, AISFormer predicts a full mask that extends beyond the ground truth, while SaVos predicts an incomplete mask. In the last image (26-11), only our \alg{} model accurately predicts the spout of the teapot.

Examining the Movi-D dataset, we note that AISFormer over-completes the predictions for the objects in the first two images (4-1 and 4-5), while SaVos delivers incomplete masks. Conversely, \alg{} accurately predicts the full masks of the books in the third (34-13) and fourth (34-19) images, while AISFormer and SaVos provide incomplete masks.

\clearpage

\subsection{Qualitative comparison in KITTI}

\begin{figure*}[!h]
\centering
\includegraphics[width=0.85\linewidth]{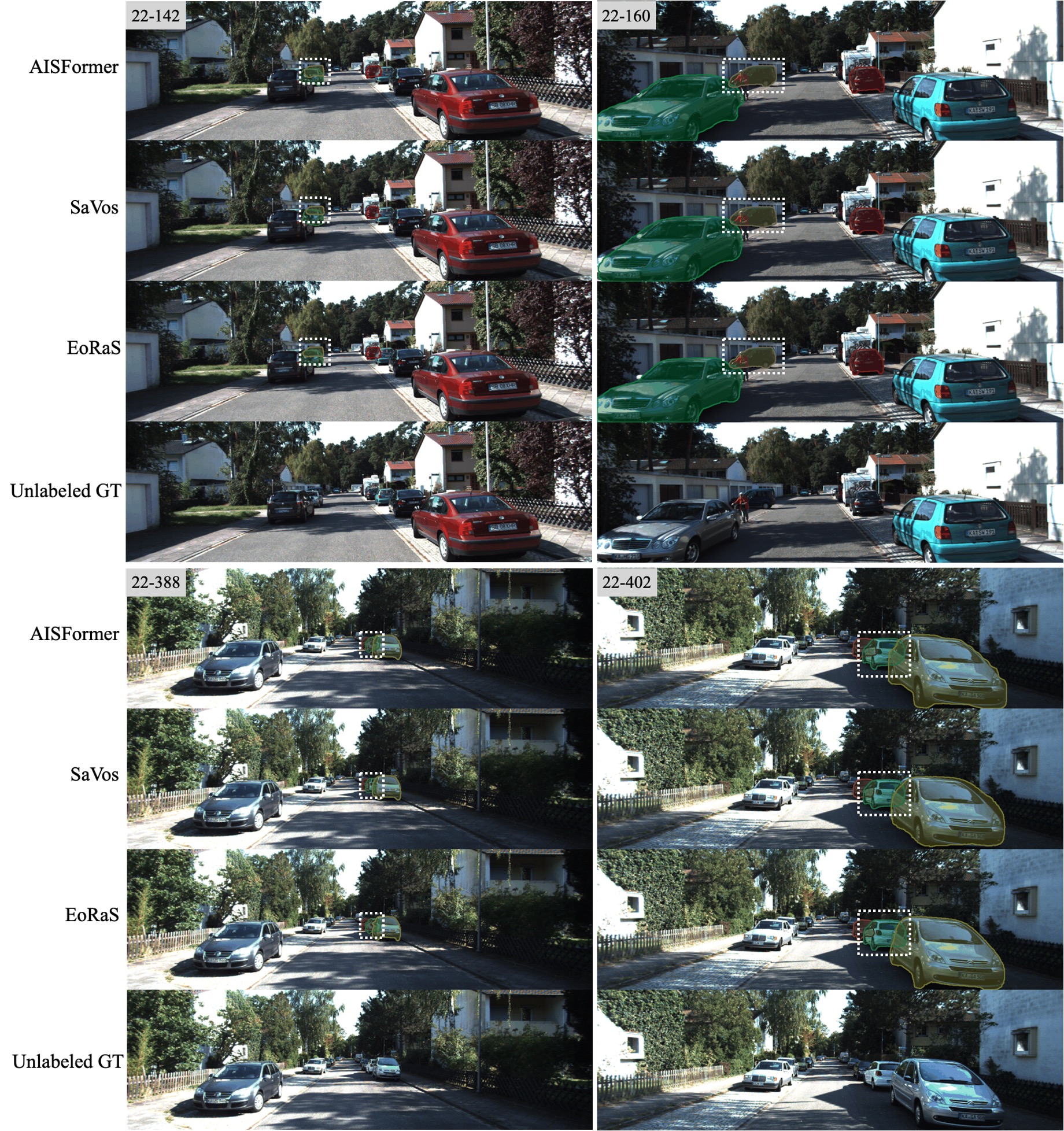}
   \vspace{-0.1in}
\caption{
Qualitative comparison between our \alg{} and competitors in the KITTI dataset. The numbers in each upper-left corner indicate where these images come from. For example, 22-142 indicates this image is from the $142^{nd}$ frame of the $22^{nd}$ video. For convenience, we also put frames on those objects with the largest predicted mask difference. Due to the sparse labeling of the KITTI dataset, many images do not have ground-truth full masks.
}
\label{fig:kitti}
\vspace{-0.2in}
\end{figure*}

In addition, we showcase the performance of our \alg{} model in the KITTI dataset (Figure \ref{fig:kitti}). Given the sparsely annotated nature of the KITTI dataset, only a few frames have annotations, with no full ground truth masks available for the selected images. Nevertheless, we observe that our model outperforms competitors in certain cases. In the upper-right image (22-160), AISFormer gives a weirdly shaped mask, while SaVos gives an over-completed mask.

To add that, we also noticed that for the cases when there is no occlusion in front of one object, \alg{} can give a more accurate mask than our competitors, as shown in the yellow mask of the lower-right image (22-402), which further shows the robustness of our model.

Overall, the results presented in Figure \ref{fig:movi} and Figure \ref{fig:kitti} suggest that our \alg{} model outperforms competitors in terms of accuracy, completeness, and robustness across various datasets.

\end{document}